\documentclass[sigconf]{acmart}
\usepackage{graphicx}
\usepackage{hyperref}
\usepackage{enumitem}
\usepackage{array} 

\newcolumntype{C}{>{\ttfamily}l} 
\usepackage{tabularx}
\usepackage{algorithm}

\setcopyright{acmlicensed}
\copyrightyear{2025}
\acmYear{2025}
\acmDOI{XXXXXXX.XXXXXXX}

\acmConference[SIGSPATIAL '26]{Make sure to enter the correct
    conference title from your rights confirmation email}{November 3-6,
    2026}{Riverside, CA}
\acmISBN{978-1-4503-XXXX-X/18/06}

\usepackage{multirow}

\begin{document}

\title{PlaceRep: Geospatial Place Representation Learning from Large-Scale Point-of-Interest Data
}
\author{Mohammad Hashemi}
\email{mohammad.hashemi@emory.edu}
\affiliation{%
    \institution{Emory University}
    \city{Atlanta}
    \country{USA}
}

\author{Hossein Amiri}
\email{hossein.amiri@emory.edu}
\affiliation{%
    \institution{Emory University}
    \city{Atlanta}
    \country{USA}
}

\author{Andreas Z{\"u}fle}
\email{azufle@emory.edu}
\affiliation{%
    \institution{Emory University}
    \city{Atlanta}
    \country{USA}
}

\renewcommand{\shortauthors}{Hashemi, et al.}
\begin{abstract}
Learning effective representations of urban environments requires capturing spatial structure beyond fixed administrative boundaries. Existing geospatial representation learning approaches typically aggregate Points of Interest (POIs) into pre-defined administrative regions such as census units or ZIP code areas, assigning a single embedding to each region. However, POIs often form semantically meaningful groups that extend across, within, or beyond these boundaries, defining \emph{places} that better reflect human activity and urban function. To address this limitation, we propose \textbf{PlaceRep}, a geospatial representation learning method that constructs place-level representations by clustering spatially and semantically related POIs. PlaceRep summarizes large-scale POI graphs from U.S. Foursquare data to produce general-purpose urban region embeddings while automatically identifying places across multiple spatial scales. By eliminating model pre-training, PlaceRep provides a scalable and efficient solution for multi-granular geospatial analysis. Experiments using the tasks of population density estimation and housing price prediction as downstream tasks show that PlaceRep outperforms most state-of-the-art graph-based geospatial representation learning methods and achieves up to a $100\times$ speedup in generating region-level representations on large-scale POI graphs. The implementation of PlaceRep is available at \url{https://github.com/mohammadhashemii/PlaceRep}.   
\end{abstract}



\maketitle

\section{Introduction}
\label{sec:introduction}
The analysis of large-scale geospatial data enables a wide range of applications that support diverse social and urban functions. A broad range of machine learning methods have been developed to learn representations from diverse geospatial data modalities, including satellite imagery, search queries, economic indicators, and mobility-related signals~\citep{du2020advances,rolf2021generalizable,ommi2024machine}. Despite their successes, existing machine learning approaches have often been limited by their task-specific nature, requiring carefully curated datasets and model architectures that generalize poorly across domains~\citep{bommasani2021opportunities}. A key step toward addressing this limitation is to identify appropriate spatial regions over which transferable geospatial representations can be learned and reused across tasks. These geospatial regions, defined as spatially distributed neighborhoods with relatively homogeneous physical and socioeconomic attributes~\citep{yuan2012discovering}, form a natural analytical unit for various urban planning tasks~\citep{zhang2017hierarchical,zhang2018integrating,cheng2022mapping}. Learning region-level embeddings from large-scale geospatial data, including POI data and satellite images, enables scalable and transferable analysis across diverse urban studies and downstream tasks~\citep{liang2025foundation}. These embeddings also serve as reusable building blocks for geospatial foundation models that require transferable representations across tasks and spatial contexts~\citep{liang2025foundation,tempelmeier2021geovectors,li2022spabert,wu2023g2ptl,hashemi2026comprehensive}.

To derive meaningful representations of geospatial regions from raw POI data, a common approach is to construct a POI graph in which nodes correspond to individual POIs and edges are defined using a geographical proximity function. General-purpose encoders are then trained on this graph to generate informative embeddings that capture the characteristics of urban regions\citep{jin2023large,yan2017itdl,agarwal2024general,huang2023learning}. 
Despite their potential, current geospatial representation learning models suffer from three key limitations that hinder their effectiveness in real-world applications:

\textbf{(a) The Missing Sense of \emph{Place} notion:}
Existing methods struggle to capture the notion of \emph{places}, regions shaped by human meaning and behavior that consist of spatially and semantically related POIs~\citep{hashemi2025points}. Most approaches learn representations at a fixed geographic granularity, assigning a single embedding to an entire region and treating it as a black box. This overlooks the fact that multiple semantically meaningful subregions may exist within the same area and may not align with administrative boundaries. For example, a ``cat lovers'' place may emerge around a cat-themed café, nearby parks frequented by stray cats, and surrounding pet shops, yet such patterns are lost under rigid spatial units. Moreover, representations based solely on individual POIs, while useful at fine spatial scales, fail to capture the broader semantic context that gives a region its meaning~\citep{niu2021delineating}.

\textbf{(b) High Pre-training Computational Cost:}
A major limitation of existing geospatial region representation methods is the high computational cost of pretraining on large POI datasets. To reduce complexity, most approaches train on sampled cities or regions, which limits scalability and realism~\citep{huang2023learning}. Extending these methods to nationwide datasets with tens of millions of POIs would incur substantial training time and resource demands.

\textbf{(c) Lack of Granularity Flexibility:}
Current geospatial region representation learning models are typically constrained to a single level of spatial granularity during both pre-training and inference time, restricting their flexibility in downstream applications. For example, PMT~\citep{wu2024pretrained} encodes trajectories as sequences of United States Census Block Groups (CBGs), with all subsequent tasks, such as next-location prediction, restricted to that level. Similarly, PDFM~\citep{agarwal2024general} produces embeddings only for ZIP codes and counties in the United States, preventing inference at finer scales such as neighborhoods or CBGs.

To address these three critical challenges, we propose \emph{PlaceRep} capable of \textbf{(1)} \textbf{Multi-granular Region Representation Learning}: generating meaningful representations for geographic entities across multiple levels of granularity, and \textbf{(2)} \textbf{Place Identification}: identifying places composed of spatially and semantically related POIs. We first define the concept of a \emph{place} and motivate its identification within POI-based geographic regions. We then introduce FSQ-19M, a large-scale POI graph dataset with approximately 19 million POIs spanning the 48 contiguous U.S. states. Building on this dataset, we propose a semantic–spatial clustering approach to jointly learn region embeddings and identify meaningful places, while preserving traceability through explicit mappings between places and their constituent POIs. Our contributions are summarized as follows:
\begin{itemize}
    \setlength{\topsep}{-0.8cm}
    \item We introduce FSQ-19M, a large-scale POI graph dataset with over 19 million POIs across the 48 contiguous U.S. states, constructed from the publicly available Foursquare dataset.  
    
    \item We formalize the notion of a \emph{place} in geospatial POI graph data and propose the first framework that automatically identifies places within a regional graph. The framework partitions POI representations into clusters, where centroids serve as place node embeddings.  
    
    \item Experiments show that PlaceRep achieves state-of-the-art accuracy and efficiency in urban representation learning on two benchmark downstream tasks: population density and housing price prediction. As shown in Figure~\ref{fig:efficiency}, PlaceRep generates ZIP code-level embeddings for the \textit{Florida} POI graph over 10$\times$ faster than existing baselines while outperforming predictive performance.  
\end{itemize}


\section{Related Works}
\label{sec:related}


Geospatial region representation learning seeks to encode geographic entities such as POIs, neighborhoods, and regions into vector embeddings that capture both functional semantics and spatial structure, enabling downstream tasks including land-use classification, recommendation, and urban analysis~\citep{zhang2024urban,liang2025foundation}. Early approaches learned POI or category embeddings based on co-occurrence patterns or spatial proximity, inspired by Word2Vec~\citep{yao2017sensing,yan2017itdl}, and were later extended to derive region-level representations by aggregating POI category embeddings~\citep{zhai2019beyond,niu2021delineating}. However, these methods often overlooked the unique spatial context of individual POIs. 

Recent work addresses this limitation by modeling POIs and regions as graph-structured data and leveraging GNNs or message-passing mechanisms to capture contextual and relational information~\citep{xu2022framework,huang2023learning}. In parallel, large-scale multimodal POI datasets have enabled pretraining geospatial foundation models that integrate signals such as satellite imagery, textual attributes, and mobility patterns~\citep{jin2023large}. Among the closest related efforts, PDFM~\citep{agarwal2024general} formulates geospatial pretraining as heterogeneous graph learning over regions enriched with diverse environmental and activity signals, learning general-purpose embeddings via GNNs. While effective for downstream tasks, PDFM operates at coarse spatial units such as ZIP codes and counties, limiting its ability to represent finer-grained urban regions such as neighborhoods.

\begin{figure}[!t]
    \vspace{-0.3cm}
    \centering
    \includegraphics[width=\columnwidth]{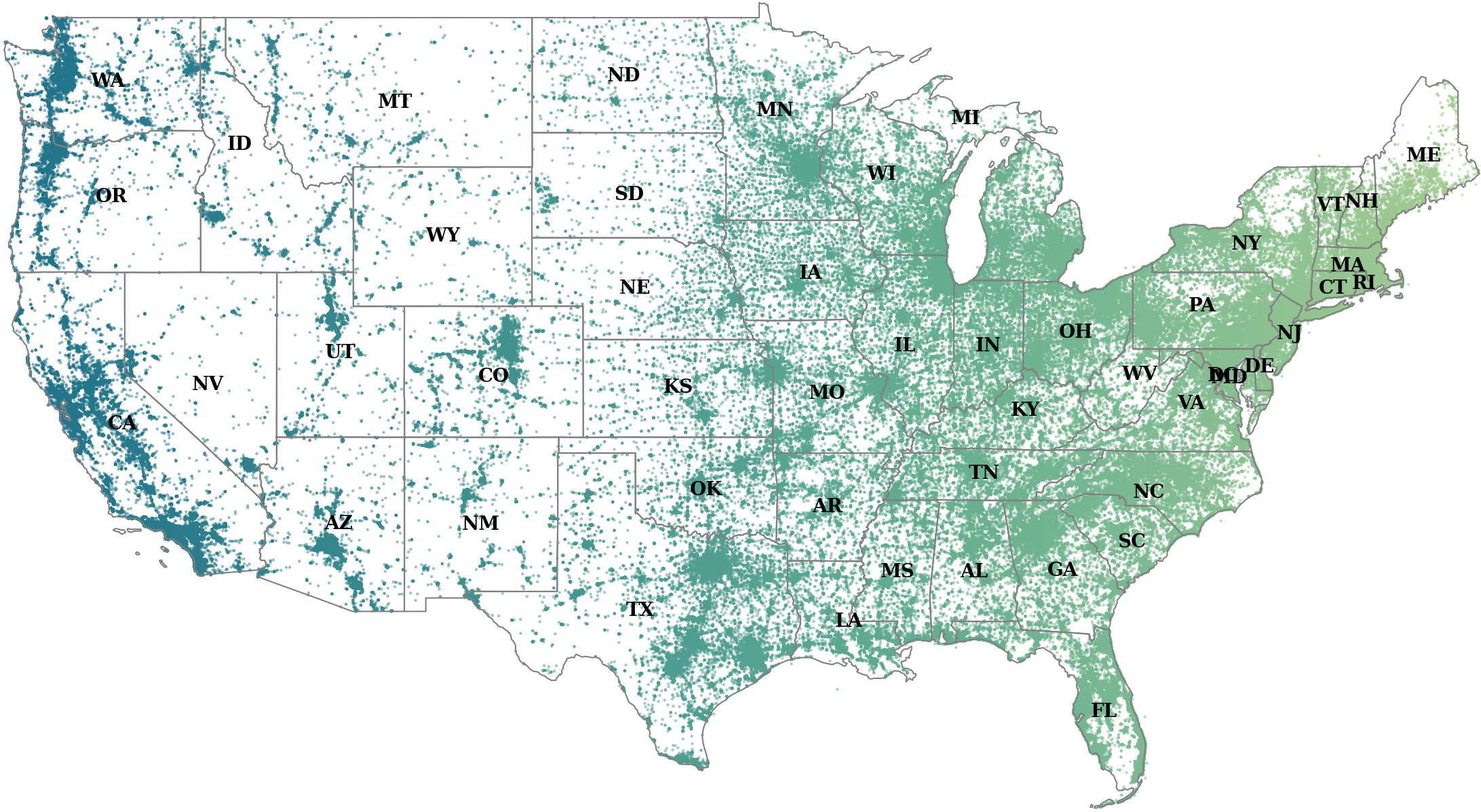}
    \caption{The spatial distribution of POIs based on a uniform random sample of one million entries from the FSQ-19M dataset, covering the 48 contiguous U.S. states.\vspace{-0.0cm}}\vspace{-0.0cm}
    \label{fig:pois}
\end{figure}

\section{Preliminaries \& Definitions}
\label{sec:preliminaries}
In this section, we present the formal definitions and problem formulation.

\subsection{\emph{Place} Definition}
\label{sec:place_definition}
Developing a geospatial representation learning method requires moving beyond point-based map representations toward a structured concept of \textit{places}~\citep{hashemi2025points}. Here, we define a place as a semantically meaningful spatial unit that may align with or encompass multiple geographic entities, including POIs, postcodes, neighborhoods, or administrative regions.

A \textit{place} $P$ is defined as a non-empty set of geographic entities:
\[
P = \{e_1, e_2, \ldots, e_n\}, \quad \text{where } e_i \in \mathcal{E}
\]
$\mathcal{E} = P \cup \mathcal{P}$ denotes the universe of geographic entities, including primitive elements $\mathcal{P}$ (i.e., POIs) and higher-level places $P$ that are both semantically and spatially similar. This recursive definition allows us to capture the hierarchical nature of places, where smaller places can be nested within larger ones. The location of each entity $e_i$ is described by their coordinates $\text{loc}_{e_i}$:\vspace{-0.4cm}

\begin{equation}
\text{loc}_{e_i} =
\begin{cases}
(\text{lat}_{p_i}, \text{lon}_{p_i}), & \text{if $e_i \in \mathcal{P}$,} \\
\text{centroid}\big(\{(\text{lat}_{p_j}, \text{lon}_{p_j}) \mid p_j \in e_i\}\big), & \text{else.}
\end{cases}
\end{equation}
%
\subsection{Problem Statement}

Let $G = (\mathcal{P}, \mathcal{E})$ denote a graph of POIs, where $\mathcal{P}$ is the set of POIs and $\mathcal{E}$ encodes their spatial relationships. We assume a predefined set of disjoint urban regions $\mathcal{U} = \{G_{r_1}, G_{r_2}, \dots, G_{r_N}\}$, where $N = |\mathcal{U}|$ denotes the number of regions and each region, i.e., a subgraph $G_{r_i} \subseteq G$ corresponds to an administrative boundary such as a neighborhood, ZIP code, county, or city. Our problem consists of two main objectives:  

\subsubsection{Region-level representation learning} The first objective is to model urban environments across multiple spatial scales by compressing large-scale POI-level data into compact and informative region-level embeddings. Formally, for each region $G_{r_i}$ we seek to learn a $d$-dimensional feature vector $z_{r_i} \in \mathbb{R}^d$ such that  $z_{r_i} = f(G_{r_i})$, where $f: \mathcal{G} \to \mathbb{R}^d$ is a representation function mapping graphs of POIs into a low-dimensional latent space. The generated embeddings $\{z_{r_i}\}_{i=1}^N$ should be both generalizable and effective across a wide range of downstream urban analytics tasks.  \vspace{-0.2cm}

\subsubsection{Place Discovery within Regions} The second objective is to automatically identify higher-level \emph{places} within each urban region $G_{r_i}$. A place $P \subseteq G_{r_i}$ as defined in Section \ref{sec:place_definition} is a subgraph consisting of POIs ro higher-level places that are both semantically and spatially similar. For each place $P$, we compute an embedding $z_{P} = f(P)$, such that $z_{P}$ captures the collective semantic and spatial characteristics of its respective POIs. This hierarchical formulation ensures traceability, since we can explicitly associate each POI with its corresponding place, and compare similarity across places both within and across regions. Such capability has applications in place discovery (e.g., identifying emerging functional areas) and place recommendation (e.g., detecting ``cat-lovers'' neighborhoods in a city).

\section{Methodology}
\label{sec:method}
\begin{figure*}[!t]
    \centering
    \includegraphics[width=\linewidth]{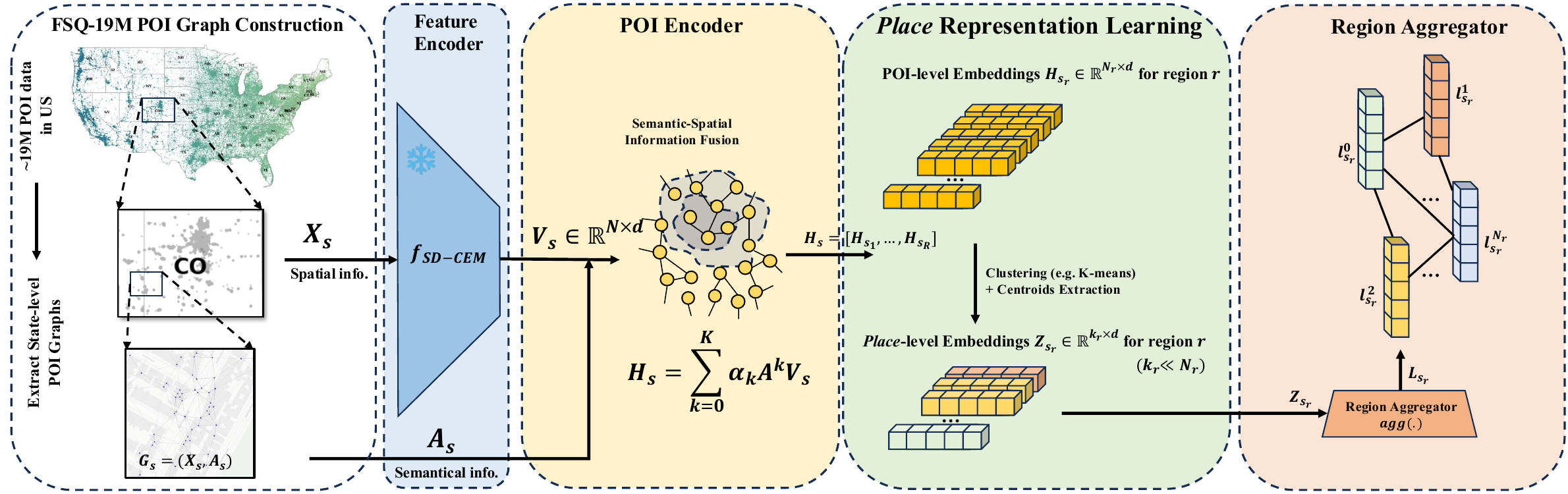}
     \vspace{-0.4cm}
    \caption{PlaceRep's architecture overview. First, it builds POI-level graphs for each state, followed by category feature encoding and feature propagation to obtain neighborhood-aware POI embeddings. Places are then identified via training-free clustering at a chosen granularity, and an aggregator function produces the final region-level embedding.\vspace{-0.2cm}}
    \label{fig:main}
\end{figure*}

In this section, we provide a detailed overview of the data preparation process, the construction of the POI-level graph, and the explanation of the gepspatial representation learning model proposed in this paper. The architecture overview of our proposed method is depicted in Figure~\ref{fig:main}.

\subsection{\emph{FSQ-19M} POI Dataset Preparation}

Our FSQ-19M dataset consists of over 19 million preprocessed and enriched POIs across the entire contiguous United States, covering all 48 states. The data is extracted from the global Foursquare POI catalog which was downloaded from the official Foursquare Places platform\footnote{\url{https://foursquare.com/products/places/}}. Foursquare POI data has been extensively used in both industry and academic research~\citep{zhao2016survey}, as it provides high-quality, frequently updated, and semantically rich representations of locations such as restaurants, stores, parks, and service providers. 

\textbf{Pre-processing pipeline:} After extracting all POIs located within the United States, we performed several pre-processing and enhancement steps to construct FSQ-19M. For each POI $p$, we retained its latitude ($\text{lat}_p$), longitude ($\text{lon}_p$), state, locality (city), and ZIP code. These attributes provide the spatial anchor for each point.

\begin{table}[ht]
\centering
\caption{A data entry in the FSQ-19M dataset.\vspace{-0.3cm}}
\label{tab:example}
\begin{tabularx}{\columnwidth}{@{}>{\ttfamily}l>{\ttfamily\arraybackslash}X@{}}
\toprule
\textbf{Attribute} & \textbf{Value} \\
\midrule
longitude & 40.858911 \\
latitude & -124.073245 \\
state & CA \\
locality & Arcata \\
postcode & 95521 \\
date\_created & 2012-02-08 \\
date\_closed & None \\
\multirow{2}{*}{category} & [ Dining and Drinking -> Restaurant -> Asian Restaurant -> Chinese Restaurant ] \\
\bottomrule
\end{tabularx}
\vspace{-0.3cm}
\end{table}

Each POI is associated with the date it was created and, if applicable, the date it was closed, enabling temporal analysis of the evolution of urban spaces. Also, each POI is assigned a hierarchical semantic category, defined up to six levels. This hierarchical structure allows for flexible aggregation at different levels of semantic granularity. We applied preprocessing steps to normalize textual attributes (e.g., consistent naming of states and ZIP codes) and to remove duplicates. Moreover, we performed a spatial join with U.S. Census ZIP code boundary data to validate and, if necessary, correct the location of each POI. The enriched dataset covers diverse urban functions, dominated by dining venues such as pizzerias, coffee shops, and fast food restaurants. Fuel stations, hair salons, churches, and fire stations are also well represented, reflecting the dataset’s commercial and civic relevance for various spatial and socioeconomic analyses.

After preprocessing, the dataset consists of \textbf{19,019,187} POIs distributed across all 48 contiguous states in the U.S., each with spatial, temporal, and semantic information. Figure~\ref{fig:pois} visualizes the spatial distribution of POIs using a uniform random sample of one million entries from FSQ-19M. Also, Table~\ref{tab:example} illustrates a representative sample from FSQ-19M, showing both spatial and semantic attributes. Section \ref{sec:graph_creation} describes the methodology used to construct a graph structure from the FSQ-19M POI dataset.

\begin{table*}[ht]
\centering
\caption{General statistics of the FSQ-19M POI graph dataset\vspace{-0.3cm}}
\label{tab:statistics}
\begin{tabular}{lrrrrr}
\toprule
\textbf{\# Total Nodes (POIs)} & 
\textbf{\# Total Edges} & 
\textbf{\# Graphs (States)} & 
\textbf{\# Total regions (ZIP codes)} & 
\textbf{\# Total sub-regions (Localities)}\\
\midrule
19{,}019{,}187 & 44{,}202{,}952 & 48 & 31{,}970 & 60{,}155\\
\bottomrule
\end{tabular}
\end{table*}



\subsection{POI Graph Data Construction}
\label{sec:graph_creation}
After obtaining the clean FSQ-19M POI-level data for the U.S., we construct 48 state-level geospatial graphs, each denoted as $G_s = (\mathbf{X}_s, \mathbf{A}_s)$, where $\mathbf{X}_s \in \mathbb{R}^{N_s \times d}$ represents the $d$-dimensional features of the $N_s$ POIs in state $s$ and $\mathbf{A}_s \in \mathbb{R}^{N_s \times N_s}$ is a weighted adjacency matrix encoding the spatial proximity between POIs. Constructing a graph is crucial because POIs with the same semantic features can have distinct functional roles depending on their surroundings. For instance, consider two coffee shops: one located in a busy shopping mall and another situated in a hospital complex. Despite sharing the same semantics, the former is likely to be associated with leisure and retail activity, whereas the latter may primarily serve healthcare visitors and staff. Capturing such contextual uniqueness is important for generating informative POI and region embeddings.

By modeling POIs as nodes in a graph, each \emph{place}, defined as a POI and its surrounding spatial context, can be represented through message passing in GCNs, enriching POI embeddings with information from neighboring POIs and capturing contextual semantics. Graph representations provide a flexible and compact structure for modeling POIs, are robust to spatial transformations, and naturally encode contextual dependencies through message passing~\citep{huang2023learning}. In this study, we adopt two widely used strategies for constructing the graph structure based on geographic proximity. 

\textbf{(1) Region-Adaptive Delaunay Triangulation} 
Delaunay triangulation (DT)~\citep{delaunay1934bulletin} is a geometric algorithm that connects a set of points in the plane such that no point lies inside the circumcircle of any triangle in the triangulation. Many previous studies have verified the fitness of DT graphs for modeling the interactions among spatial vector data: this method ensures that edges connect spatially close POIs while avoiding overly dense connections, thereby preserving the local neighborhood structure while maintaining computational efficiency~\citep{huang2023learning,xu2022framework,yan2019graph,huang2022estimating}.  

For each pair of POIs \(i\) and \(j\) connected by the triangulation, we assign a weighted edge in the adjacency matrix \(\mathbf{A}_s\) defined as  \vspace{-0.18cm}
\[
A_{ij} = \log \left( \frac{1 + L_r^{1.5}}{1 + d(i,j)^{1.5}} \right) \cdot w_{r}(i,j),
\]
where \(d(i,j)\) denotes the geographic distance between POIs \(i\) and \(j\), which can be computed either as Euclidean distance in projected coordinates or Haversine distance on the sphere. 

Here, \(L_r\) is a region-specific scaling factor, computed using the POIs within the subgraph \(G_r \subseteq G_s\) corresponding to a local region of POIs. This local, adaptive scaling is important because the spatial density of POIs can vary drastically even within the same state. 
By computing \(L_r\) for each region \(G_r\), the weighting function remains balanced, ensuring that the edge weights are comparable across areas with highly variable local densities. Following~\citep{huang2023learning}, the factor \(w_{r}(i,j)\) encodes regional consistency: it is set to \(1.0\) if POIs \(i\) and \(j\) belong to the same region, and to \(0.4\) if they connect across different regions, thereby down-weighting cross-region connections. The computed edge weights are then scaled uniformly within each region to the range \([0, 1]\), resulting in \(\mathbf{A}_s\) being a normalized weighted adjacency matrix.

\textbf{(2) k-Nearest Neighbors (KNN).} For each POI \(i\), we connect its \(k\) closest neighbors based on geographic distance (Euclidean or Haversine) in the adjacency matrix \(\mathbf{A}_s\). Edges are unweighted (0 or 1), indicating the presence of a connection. Despite its simplicity, this approach balances neighbor consistency and spatial proximity: each node connects to its nearest neighbors, capturing local interactions while preventing overly dense graphs in high-density areas.


\subsection{\emph{Place} Representation Generation}

PlaceRep is a training-free method designed to generate meaningful urban region embeddings from graph-based POI data and to identify places within a given geographical region. The overall architecture of PlaceRep is shown in Figure~\ref{fig:main}. In this section, we provide a detailed description of each component of the framework:

\subsubsection{Feature Encoder}

PlaceRep enables encoding POI-level raw attributes to extract rich and meaningful features, facilitating effective representation learning for a variety of urban-related tasks. In this study, we leverage the semantic category of each POI as the primary attribute, which is inherently hierarchical, capturing fine-grained distinctions between different types of POIs. To model these hierarchical category attributes, we utilize SD-CEM~\citep{jia2024learning}, a semantically disentangled POI category embedding model. It generates hierarchy-enhanced category representations by pre-training on large-scale mobility sequences, learning disentangled embeddings that capture semantic relationships among POI categories.

Formally, let $\mathcal{C}$ denote the set of POI categories with $L$ hierarchical levels. For a POI $p$, its category labels across levels are denoted by $\{c_1, c_2, \dots, c_L\}$, where $c_1$ is the most general level and $c_L$ the most specific. Formally. given a POI category $c_i$, it is mapped to a vector representation $\mathbf{v}_{c_i} \in \mathbb{R}^d$, such that
\[
\mathbf{v}_p = \text{SD-CEM}(\{c_1, \dots, c_L\}) \in \mathbb{R}^d,
\]
where $\mathbf{v}_p$ captures both the hierarchical and semantic information of the POI.

\begin{table*}[ht!]
\vspace{-0.3cm}
\centering
\caption{ZIP-code level prediction performance for population density \(\left(\text{people} / \text{km}^2\right)\). 
The best results are highlighted in \textbf{bold}, and values in parentheses indicate the standard deviation.\vspace{-0.3cm}
}

\label{tab:main_pd}
\resizebox{\textwidth}{!}{%
\begin{tabular}{cccccccccccccccc}
\toprule
\multirow{2}{*}{\textbf{State (\# POIs)}} 
& \multicolumn{3}{c}{\textbf{Averaging}}  
& \multicolumn{3}{c}{\textbf{Place2Vec~\citep{yan2017itdl}}}  
& \multicolumn{3}{c}{\textbf{HGI~\citep{huang2023learning}}}  
& \multicolumn{3}{c}{\textbf{PDFM~\citep{agarwal2024general}}}  
& \multicolumn{3}{c}{\textbf{PlaceRep (ours)}} \\ 
\cmidrule(l){2-4}\cmidrule(l){5-7}\cmidrule(l){8-10}\cmidrule(l){11-13}\cmidrule(l){14-16}
& RMSE $\downarrow$ & MAE $\downarrow$ & $R^2$ $\uparrow$ &
RMSE $\downarrow$ & MAE $\downarrow$ & $R^2$ $\uparrow$ &
RMSE $\downarrow$ & MAE $\downarrow$ & $R^2$ $\uparrow$ &
RMSE $\downarrow$ & MAE $\downarrow$ & $R^2$ $\uparrow$ &
RMSE $\downarrow$ & MAE $\downarrow$ & $R^2$ $\uparrow$ \\
\midrule

\multirow{2}{*}{\textit{WY} ($\sim40K$)} 
  & 65.67 & 22.44 & -5.71 
  & 62.12 & 21.14 & -1.32 
  & 69.52 & 19.66 & -9.51 
  & 90.39 & 30.22 & -13.56 
  & \textbf{61.14} & \textbf{18.14} & \textbf{-1.27} 
\\
  & ($\pm$43.86) & ($\pm$9.38) & ($\pm$9.63) 
  & ($\pm$32.22) & ($\pm$18.19) & ($\pm$18.01) 
  & ($\pm$47.14) & ($\pm$11.38) & ($\pm$21.06) 
  & ($\pm$44.05) & ($\pm$8.58) & ($\pm$25.50) 
  & \textbf{($\pm$38.00)} & \textbf{($\pm$7.55)} & \textbf{($\pm$22.87)} 
\\ 

\multirow{2}{*}{\textit{VT} ($\sim52K$)}
  & 265.38 & 95.08 & -1.87 
  & 240.76 & 78.08 & -1.72 
  & 170.94 & \textbf{54.76} & -1.33 
  & 308.18 & 107.07 & -2.47 
  & \textbf{165.23} & 56.82 & \textbf{-0.89} 
\\ 
  & ($\pm$146.43) & ($\pm$38.75) & ($\pm$36.18) 
  & ($\pm$87.11) & ($\pm$22.13) & ($\pm$21.12) 
  & ($\pm$112.29) & \textbf{($\pm$30.33)} & ($\pm$3.81) 
  & ($\pm$237.26) & ($\pm$32.21) & ($\pm$2.78) 
  & \textbf{($\pm$88.75)} & ($\pm$28.19) & \textbf{($\pm$4.65)} 
\\ 
\midrule

\multirow{2}{*}{\textit{AL ($\sim255K$)}} 
  & 283.72 & 144.86 & 0.02 
  & 290.11 & 144.76 & 0.02 
  & 280.57 & 113.87 & 0.03 
  & 268.16 & 115.43 & 0.16 
  & \textbf{266.24} & \textbf{140.44} & \textbf{0.25} 
\\
 
  & ($\pm$27.90) & ($\pm$12.13) & ($\pm$0.48) 
  & ($\pm$22.32) & ($\pm$16.12) & ($\pm$0.29) 
  & ($\pm$46.08) & ($\pm$10.77) & ($\pm$0.62) 
  & ($\pm$68.00) & ($\pm$11.01) & ($\pm$0.17) 
  & \textbf{($\pm$18.83)} & \textbf{($\pm$9.31)} & \textbf{($\pm$0.56)} 
\\ 

\multirow{2}{*}{\textit{GA ($\sim652K$)}} 
  & 532.00 & 197.31 & 0.39 
  & 511.21 & 187.14 & 0.47 
  & 458.21 & 167.16 & 0.53 
  & 536.34 & 186.33 & 0.38 
  & \textbf{410.16} & \textbf{155.02} & \textbf{0.89} 
\\  
  & ($\pm$139.75) & ($\pm$13.60) & ($\pm$0.22) 
  & ($\pm$201.25) & ($\pm$25.10) & ($\pm$0.14) 
  & ($\pm$171.65) & ($\pm$25.52) & ($\pm$0.26) 
  & ($\pm$155.65) & ($\pm$25.96) & ($\pm$0.17) 
  & \textbf{($\pm$112.21)} & \textbf{($\pm$26.16)} & \textbf{($\pm$0.31)} 
\\ 
  
\midrule

\multirow{2}{*}{\textit{NY ($\sim1.24M$)}} 
  & 4872.13 & 1841.26 & 0.19 
  & 4644.87 & 1712.16 & 0.21 
  & \textbf{4305.66} & \textbf{1468.68} & \textbf{0.61} 
  & 4483.95 & 1507.88 & 0.44 
  & 4410.86 & 1497.12 & 0.52 
\\ 
  & ($\pm$722.78) & ($\pm$246.76) & ($\pm$0.10) 
  & ($\pm$589.28) & ($\pm$126.06) & ($\pm$0.12) 
  & \textbf{($\pm$843.69)} & \textbf{($\pm$270.21)} & \textbf{($\pm$0.09)} 
  & ($\pm$700.85) & ($\pm$300.87) & ($\pm$0.11) 
  & ($\pm$642.54) & ($\pm$318.22) & ($\pm$0.12) 
\\ 

\multirow{2}{*}{\textit{FL ($\sim1.35M$)}} 
  & 1035.25 & 526.33 & 0.244 
  & 1115.62 & 615.21 & 0.16 
  & 772.30 & 351.21 & 0.50 
  & 806.30 & 373.98 & 0.35 
  & \textbf{702.56} & \textbf{310.72} & \textbf{0.56} 
\\
  & ($\pm$297.92) & ($\pm$54.55) & ($\pm$0.25) 
  & ($\pm$410.65) & ($\pm$76.11) & ($\pm$0.31) 
  & ($\pm$310.42) & ($\pm$82.12) & ($\pm$0.21) 
  & ($\pm$257.88) & ($\pm$37.63) & ($\pm$0.09) 
  & \textbf{($\pm$101.75)} & \textbf{($\pm$23.21)} & \textbf{($\pm$0.10)} 
\\ 

\multirow{2}{*}{\textit{CA ($\sim2.2M$)}}  
  & 1665.47 & 963.78 & 0.40 
  & 1640.90 & 950.16 & 0.44 
  & 1540.18 & 819.69 & 0.50 
  & 1269.25 & 665.70 & 0.66 
  & \textbf{1053.43} & \textbf{588.14} & \textbf{0.69} 
\\
 
  & ($\pm$170.76) & ($\pm$351.29) & ($\pm$0.10) 
  & ($\pm$321.71) & ($\pm$244.11) & ($\pm$0.11) 
  & ($\pm$235.74) & ($\pm$58.58) & ($\pm$0.07) 
  & ($\pm$178.21) & ($\pm$40.35) & ($\pm$0.05) 
  & \textbf{($\pm$152.01)} & \textbf{($\pm$77.54)} & \textbf{($\pm$0.06)} 
\\ 

\bottomrule
\end{tabular}}
\end{table*}

\subsubsection{POI Encoder}
Upon encoding the features of each POI into a latent $d$-dimensional representation $\mathbf{v}_p$, we further need to enrich the representation of the POI itself. While the previous step captures the semantics of POI categories, it overlooks the uniqueness of each individual POI. In practice, POIs with the same semantic label can exhibit distinct characteristics depending on their surrounding environment. Intuitively, the uniqueness of a POI is shaped by its spatial context. 

To capture the contextual semantics of \textit{places}, it is therefore essential to move beyond isolated POI attributes and incorporate region-level structural information. For instance, a Starbucks located near university buildings and departments carries very different contextual meaning compared to one situated along a remote highway rest stop. To this end, PlaceRep employs a lightweight, non-parametric graph propagation mechanism that enriches each POI’s representation by incorporating information from its spatial neighbors.

Formally, let $G = (\mathcal{P}, \mathcal{E})$ denote the POI graph, where $\mathcal{P}$ is the set of POIs and $\mathcal{E}$ the spatial edges connecting nearby POIs. Each node $p \in \mathcal{P}$ is initially associated with a feature vector $\mathbf{v}_p \in \mathbb{R}^d$. We follow the propagation method of Simplified Graph Convolution (SGC)~\citep{wu2019simplifying}, where the propagated representations after $k$ steps are computed as:\vspace{-0.15cm}
\begin{equation}
    \mathbf{H}^{(k)} = \hat{\mathbf{A}}^k \mathbf{V}, \quad \mathbf{V} \in \mathbb{R}^{|\mathcal{V}| \times d},
\end{equation}
where $\hat{\mathbf{A}}$ is the symmetrically normalized adjacency matrix of $G$, defined as $\hat{\mathbf{A}} = \tilde{\mathbf{D}}^{-\frac{1}{2}} \tilde{\mathbf{A}} \tilde{\mathbf{D}}^{-\frac{1}{2}}$.

with $\tilde{\mathbf{A}} = \mathbf{A} + \mathbf{I}$ being the adjacency matrix with self-loops, and $\tilde{\mathbf{D}}$ the degree matrix of $\tilde{\mathbf{A}}$ such that $(\tilde{\mathbf{D}})_{ii} = \sum_j \tilde{\mathbf{A}}_{ij}$.

To aggregate information from multiple propagation depths, we adopt a weighted linear combination across steps:
\begin{equation}
    \mathbf{H} = \sum_{k=0}^{K} \alpha_k \mathbf{H}^{(k)},
    \label{equ:poi_encoder}
\end{equation}
where $\alpha_k \in \mathbb{R}$ is the propagation weight at step $k$. This formulation controls the extent to which semantic and spatial information is propagated, thereby allowing PlaceRep to act as a spatially and semantically controlled approach. In doing so, it ensures that each POI’s embedding integrates both local and multi-hop neighborhood semantics. Importantly, we allow $\alpha_k$ to vary across a wide range, including negative values. When $\alpha_k < 0$, the model effectively introduces a ``negative offset'' that captures heterophilic relationships between POIs~\citep{zhu2021graph}, which is particularly useful in urban settings where co-located POIs may have contrasting functions (e.g., a bar next to a church).

\subsubsection{Place Representation Learning via Clustering}
Once we obtain the POI embeddings, the next step is to generate region-level embeddings and to identify places within a region, as introduced as the two main objectives of PlaceRep in Section~\ref{sec:preliminaries}. Unlike existing urban region representation learning methods, which are typically restricted to a fixed spatial resolution, PlaceRep is inherently multigranular, enabling the generation of embeddings at arbitrary levels of granularity. This flexibility accommodates both geographic zones and non-geographic groupings (e.g., functional regions, semantic clusters). 

From a data-centric AI perspective~\citep{zha2025data}, one effective approach to summarizing a large-scale POI graph while preserving its semantic and spatial structure is to learn representative embeddings for aggregated POIs within each region. This process, known as \textit{graph reduction}~\citep{hashemi2024comprehensive}, reduces the size of the graph while retaining sufficient expressivity for downstream tasks. Inspired by GECC~\citep{gong2025scalable}, a clustering-based graph reduction method that theoretically guarantees comparable performance for GNNs trained on condensed graphs, PlaceRep introduces an urban representation learning component that leverages clustering over propagated POI embeddings to construct semantically meaningful and context-aware places. This transforms fine-grained POI embeddings into higher-level \textit{place} representations.

Formally, let $\mathbf{H} \in \mathbb{R}^{n \times d}$ denote the matrix of POI embeddings for a region $r$, where $n$ is the number of POIs and $d$ the embedding dimension. For each predefined region $G_r = (\mathcal{V}_r, \mathcal{E}_r)$, we extract its embeddings:
\begin{equation}
    \mathbf{H}_r = \{\mathbf{h}_p \mid p \in \mathcal{V}_r\}, \quad \mathbf{H}_r \in \mathbb{R}^{n_r \times d},
\end{equation}
where $n_r = |\mathcal{V}_r|$ is the number of POIs in region $r$. 

To extract \textit{places} (i.e., sub-regions of semantically and spatially similar POIs), we apply bisecting $k$-means clustering~\citep{rohilla2019data}. Unlike standard $k$-means, bisecting $k$-means iteratively splits the largest cluster into two sub-clusters until the desired number of clusters $k_r$ is reached, naturally producing a hierarchical structure that aligns with the hierarchical semantics of POI embeddings. The optimization objective is given by:
\begin{equation}
    \min_{\{\mathcal{C}_1, \dots, \mathcal{C}_{k_r}\}} \sum_{j=1}^{k_r} \sum_{\mathbf{h}_p \in \mathcal{C}_j} \|\mathbf{h}_p - \boldsymbol{\mu}_j\|_2^2,
\end{equation}
where $\boldsymbol{\mu}_j = \frac{1}{|\mathcal{C}_j|} \sum_{\mathbf{h}_p \in \mathcal{C}_j} \mathbf{h}_p$ is the centroid of cluster $\mathcal{C}_j$.

We introduce a reduction ratio hyperparameter $r \in [0, 1]$ to control the number of places per region: $k_r = \left\lfloor n_r \times r \right\rfloor$, where smaller values of $r$ yield fewer but coarser places, and larger values of $r$ produce finer-grained places. This formulation allows flexible trade-offs between representational detail and computational efficiency. In cases where the distribution of POIs is highly non-uniform, density-based clustering methods such as DBSCAN~\citep{ester1996density} can also be employed, which do not require specifying $k_r$ in advance.

The resulting set of clusters $\{\mathcal{C}_1, \dots, \mathcal{C}_{k_r}\}$ serves as the \textit{places} representing region $r$. Each cluster is summarized by its centroid embedding: $\mathbf{z}_j = \frac{1}{|\mathcal{C}_j|} \sum_{\mathbf{h}_p \in \mathcal{C}_j} \mathbf{h}_p$, where 
\begin{equation}
    \mathbf{z}_j = \frac{1}{|\mathcal{C}_j|} \sum_{\mathbf{h}_p \in \mathcal{C}_j} \mathbf{h}_p, \quad j=1,\dots,k_r,
\end{equation}
yielding a condensed representation $\mathbf{Z}_r = \{\mathbf{z}_1, \dots, \mathbf{z}_{k_r}\} \in \mathbb{R}^{k_r \times d}$. It has been theoretically proven~\citep{gong2025scalable} that such condensed representations are as expressive as the original POI embeddings, allowing downstream models trained on $\mathbf{Z}_r$ to achieve comparable performance to those trained on $\mathbf{H}_r$, while the size of $\mathbf{Z}_r$ is significantly smaller than $\mathbf{H}_r$, resulting in much higher efficiency during model training.

\subsubsection{Region Aggregator}
Once the place embeddings $\mathbf{Z}_r$ are obtained, in order to generate a single meaningful representation for the entire region $r$, a simple and efficient aggregation function $\text{agg}$ is performed, which computes the weighted average of all place embeddings, weighted by the number of POIs in each place. Formally, the aggregated regional embedding is given by:
    $\mathbf{L}_r = \big[ \, \mathbf{l}_1 \; | \; \mathbf{l}_2 \; | \; \dots \; | \; \mathbf{l}_{N_r} \, \big]$, where each $\mathbf{l}_j$ is a place embedding computed as the weighted average of POIs in place $\mathbf{l}_j = \frac{\sum_{i=1}^{N_j} n_i \cdot \mathbf{z}_i}{\sum_{i=1}^{N_j} n_i}$, where $\mathbf{z}_i$ is the embedding of place $i$, $n_i$ is the number of POIs contained in place $i$, and $N_r$ is the total number of places in region $r$. An adjacency matrix is constructed to represent the connectivity between region embeddings, where two regions are considered connected if their polygon boundaries touch each other.







\vspace{-0.5cm}
\section{Experiments}
\label{sec:experiments}

\subsection{Experimental Setup}

\begin{table*}[ht!]
\centering
\caption{ZIP-code level Prediction performance for the housing price using the Zillow dataset for August 2024 (ZHVI/$10^3$). The best results are highlighted in \textbf{bold}, and values in parentheses indicate the standard deviation.\vspace{-0.3cm}
}

\label{tab:main_hp}
\resizebox{\textwidth}{!}{%
\begin{tabular}{ccccccccccccccccc}
\toprule
\multirow{2}{*}{\textbf{State (\# POIs)}} 
& \multicolumn{3}{c}{\textbf{Averaging}}  
& \multicolumn{3}{c}{\textbf{Place2Vec~\citep{yan2017itdl}}}  
& \multicolumn{3}{c}{\textbf{HGI~\citep{huang2023learning}}}  
& \multicolumn{3}{c}{\textbf{PDFM~\citep{agarwal2024general}}}  
& \multicolumn{3}{c}{\textbf{PlaceRep (ours)}} \\ 
\cmidrule(l){2-4}\cmidrule(l){5-7}\cmidrule(l){8-10}\cmidrule(l){11-13}\cmidrule(l){14-16}
& RMSE $\downarrow$ & MAE $\downarrow$ & $R^2$ $\uparrow$ &
RMSE $\downarrow$ & MAE $\downarrow$ & $R^2$ $\uparrow$ &
RMSE $\downarrow$ & MAE $\downarrow$ & $R^2$ $\uparrow$ &
RMSE $\downarrow$ & MAE $\downarrow$ & $R^2$ $\uparrow$ &
RMSE $\downarrow$ & MAE $\downarrow$ & $R^2$ $\uparrow$ \\
\midrule

\multirow{2}{*}{\textit{WY} ($\sim40K$)} 
  & 370.45 & 217.81 & -3.26 
  & 392.82 & 254.92 & -4.02 
  & 361.39 & 220.70 & -3.01 
  & 388.68 & 223.10 & -3.27 
  & \textbf{350.21} & \textbf{202.88} & \textbf{-2.87} 
\\
  & ($\pm$111.08) & ($\pm$551.76) & ($\pm$5.23) 
  & ($\pm$122.14) & ($\pm$612.78) & ($\pm$2.01) 
  & ($\pm$152.06) & ($\pm$689.04) & ($\pm$1.90) 
  & ($\pm$145.45) & ($\pm$565.10) & ($\pm$4.52) 
  & \textbf{($\pm$122.10)} & \textbf{($\pm$341.21)} & \textbf{($\pm$2.33)} 
\\ 

\multirow{2}{*}{\textit{VT} ($\sim52K$)}
  & 138.87 & 96.43 & -0.20 
  & 144.02 & 111.38 & -0.28 
  & 139.29 & 100.48 & -0.23 
  & 210.18 & 157.12 & -2.02 
  & \textbf{135.66} & \textbf{93.83} & \textbf{-0.16} 
\\ 
  & ($\pm$239.49) & ($\pm$138.08) & ($\pm$0.15) 
  & ($\pm$128.01) & ($\pm$25.15) & ($\pm$0.21) 
  & ($\pm$20.05) & ($\pm$9.38) & ($\pm$0.17) 
  & ($\pm$189.77) & ($\pm$48.11) & ($\pm$0.98) 
  & \textbf{($\pm$22.16)} & \textbf{($\pm$13.16)} & \textbf{($\pm$0.21)} 
\\ 
\midrule

\multirow{2}{*}{\textit{AL ($\sim255K$)}} 
  & 96.63 & 73.74 & -0.17 
  & 96.57 & 72.16 & -0.16 
  & 97.55 & 73.23 & -0.20 
  & 96.14 & 72.38 & -0.18 
  & \textbf{89.85} & \textbf{66.93} & \textbf{-0.02} 
\\
 
  & ($\pm$10.74) & ($\pm$73.74) & ($\pm$0.08) 
  & ($\pm$15.14) & ($\pm$9.14) & ($\pm$0.19) 
  & ($\pm$11.00) & ($\pm$5.16) & ($\pm$0.11) 
  & ($\pm$8.51) & ($\pm$3.38) & ($\pm$0.17) 
  & \textbf{($\pm$10.63)} & \textbf{($\pm$3.76)} & \textbf{($\pm$0.02)} 
\\ 

\multirow{2}{*}{\textit{GA ($\sim652K$)}} 
  & 186.53 & 125.82 & -0.24 
  & 192.01 & 131.20 & -0.28 
  & 186.96 & 122.59 & -0.25 
  & 197.52 & 118.50 & -0.52 
  & \textbf{171.04} & \textbf{115.33} & \textbf{-0.02} 
\\  
  & ($\pm$67.55) & ($\pm$9.74) & ($\pm$0.21) 
  & ($\pm$78.11) & ($\pm$15.78) & ($\pm$0.16) 
  & ($\pm$70.09) & ($\pm$10.66) & ($\pm$0.24) 
  & ($\pm$90.25) & ($\pm$17.15) & ($\pm$1.10) 
  & \textbf{($\pm$70.66)} & \textbf{($\pm$9.31)} & \textbf{($\pm$0.02)} 
\\ 
  
\midrule

\multirow{2}{*}{\textit{NY ($\sim1.24M$)}} 
  & 423.59 & 278.70 & -0.10 
  & 421.65 & 270.21 & -0.09 
  & 414.53 & 267.44 & -0.06 
  & \textbf{410.11} & \textbf{251.08} & \textbf{-0.02} 
  & 418.94 & 275.29 & -0.07 
\\ 
  & ($\pm$49.718) & ($\pm$114.84) & ($\pm$0.08) 
  & ($\pm$60.70) & ($\pm$11.23) & ($\pm$0.11) 
  & ($\pm$53.92) & ($\pm$15.42) & ($\pm$0.07) 
  & \textbf{($\pm$49.04)} & \textbf{($\pm$11.21)} & \textbf{($\pm$0.12)} 
  & ($\pm$50.77) & ($\pm$12.45) & ($\pm$0.05) 
\\ 

\multirow{2}{*}{\textit{FL ($\sim1.35M$)}} 
  & 362.11 & 185.29 & -0.20 
  & 382.28 & 201.07 & -0.39 
  & 370.78 & 182.53 & -0.19 
  & 375.30 & 189.67 & -0.34 
  & \textbf{338.39} & \textbf{169.20} & \textbf{-0.01} 
\\
  & ($\pm$91.30) & ($\pm$15.14) & ($\pm$0.21) 
  & ($\pm$110.65) & ($\pm$16.11) & ($\pm$0.31) 
  & ($\pm$87.71) & ($\pm$15.39) & ($\pm$0.20) 
  & ($\pm$86.26) & ($\pm$12.14) & ($\pm$0.38) 
  & \textbf{($\pm$104.42)} & \textbf{($\pm$12.09)} & \textbf{($\pm$0.01)} 
\\ 

\multirow{2}{*}{\textit{CA ($\sim2.2M$)}}  
  & 692.38 & 473.29 & -0.08 
  & 688.21 & 467.00 & -0.07 
  & 685.07 & 461.32 & -0.06 
  & 701.09 & 471.23 & -0.11 
  & \textbf{669.69} & \textbf{440.19} & \textbf{-0.01} 
\\
 
  & ($\pm$687.57) & ($\pm$25.32) & ($\pm$0.05) 
  & ($\pm$78.20) & ($\pm$24.89) & ($\pm$0.07) 
  & ($\pm$64.01) & ($\pm$20.28) & ($\pm$0.06) 
  & ($\pm$66.74) & ($\pm$25.38) & ($\pm$0.06) 
  & \textbf{($\pm$76.03)} & \textbf{($\pm$27.87)} & \textbf{($\pm$0.01)} 
\\ 

\bottomrule
\end{tabular}}
\end{table*}

The FSQ-19M POI graph dataset is constructed as a collection of 48 homogeneous graphs, each corresponding to one of the contiguous U.S. states. Splitting the dataset at the state level allows us to evaluate PlaceRep across graphs with diverse densities and structural statistics. For instance, Wyoming (WY) forms the smallest graph, containing 40{,}165 POIs, while California (CA) constitutes the largest, with 2{,}204{,}300 POIs. This setup provides a natural testbed for assessing the scalability and robustness of PlaceRep in comparison with existing methods under varying graph sizes and complexities.

We benchmark PlaceRep against several representative baselines, ranging from simple heuristics to state-of-the-art geospatial foundation models, all of which generate region-level embeddings within each state. For a fair comparison, the dimensionality of the generated region embeddings is fixed at $ d = 30 $ across all models.

1. \textbf{Averaging:} A simple heuristic method in which the embedding of each region is computed by averaging the embeddings of its POI categories.

2. \textbf{Place2Vec~\citep{yan2017itdl}:} A method that learns POI category embeddings and then derives region embeddings by averaging over the POIs contained in a region. Unlike simple averaging, Place2Vec incorporates spatial co-occurrence statistics during POI embedding learning, yet the final regional representation is still obtained through aggregation by averaging.

3. \textbf{HGI~\citep{huang2023learning}:} An unsupervised model that learns region embeddings by jointly modeling categorical semantics of POIs, POI-level and region-level adjacency, and the interaction between POIs and regions. HGI further employs an attention mechanism to weigh the relative importance of individual POIs during region-level aggregation, enabling more context-sensitive representations.

4. \textbf{PDFM~\citep{agarwal2024general}:} A pre-trained foundation model that integrates diverse geospatial signals, including POI data, maps, activity levels, search trends, weather, and air quality, into a heterogeneous graph. Geographic regions such as counties and postal codes are represented as nodes, with edges reflecting spatial proximity. A GNN is then applied to capture spatial dependencies and generate embeddings for these regions.

\subsubsection{Implementation Details}

To ensure a fair reproduction and comparison with baseline methods, we reimplemented all baselines and tuned their hyperparameters, guided both by the best configurations reported in the original papers and by additional fine-tuning under our experimental setting. The complete implementation, including code for PlaceRep and all baseline models, is publicly available in the PlaceRep repository\footnote{\url{https://github.com/mohammadhashemii/PlaceRep}}. To maintain consistency, the number of evaluations for downstream tasks was restricted to 10. All baselines, with the exception of PDFM~\citep{agarwal2024general}, were trained from scratch. For PDFM, due to the unavailability of detailed architectural specifications, we reused the released pretrained embeddings. Since PDFM integrates heterogeneous data sources (e.g., maps, activity levels, search trends, weather, and air quality), we only utilized the map-based embeddings and further applied dimensionality reduction to match the embedding size to $d = 30$, ensuring a relatively fair comparison with other methods.

For PlaceRep, we determined that the optimal dimensionality of the feature encoder was $d=30$. The POI encoder was implemented using a single-layer SGC~\citep{wu2019simplifying} with propagation limited to two steps (capturing up to second-order neighborhood information). The propagation coefficients were tuned within the range $\alpha_0, \alpha_1, \alpha_2 \in [0.0, 1.0]$ with increments of $0.25$. For the clustering module, the maximum number of iterations for $k$-means was set to 300, with a convergence threshold of $1 \times 10^{-8}$. Each experiment was repeated 10 times, and we report the averaged results to ensure statistical reliability.

To efficiently execute the clustering procedure, we relied on Intel(R) Xeon(R) Platinum 8260 CPUs @ 2.40GHz with NumPy~\citep{harris2020array} for numerical computation. Baseline model training was conducted on a high-performance computing cluster equipped with a heterogeneous mix of GPUs: Tesla A100 (40GB) and V100 (32GB) for large-scale graphs, and K80 (12GB) GPUs for smaller graphs. This heterogeneous setup enabled efficient handling of datasets of varying sizes while ensuring consistent evaluation across all methods.

\subsection{Downstream tasks performance analysis}
\label{sec:downstream_tasks}

\subsubsection{Population Density Prediction}
Understanding the spatial distribution of human populations is fundamental to a wide range of operational tasks, policy design, and scientific research, including disaster response, infrastructure development, and urban planning~\citep{huang2023learning}. Conventional census-based approaches to collecting population data, while reliable, are both labor-intensive and limited in spatiotemporal resolution. Consequently, recent studies have increasingly turned to alternative geospatial data sources, such as remote sensing imagery and POIs, to enable fine-grained population estimation~\citep{yao2017mapping,shang2021estimating}.

In this experiment, we use ZIP-code level region representations generated by PlaceRep to estimate population density, enabling direct comparison with baseline methods. Following prior work, we train a Random Forest regressor with 100 decision trees using an 80/20 train-test split over ZIP-code embeddings. Population density labels are obtained from publicly available U.S. Census Bureau datasets.
Each experiment is repeated 10 times with randomized train-test splits, and the average performance is reported. Results are summarized in Table~\ref{tab:main_pd}. For fairness, all baseline models are carefully reimplemented and tuned using their optimal hyperparameter configurations; these settings are reported in Table~\ref{tab:hyperparam} for reproducibility.

Empirical results demonstrate that PlaceRep consistently outperforms baseline approaches across nearly all datasets, underscoring the effectiveness of our training-free region representation learning paradigm. The only exception arises in New York, where PlaceRep ranks second by a marginal gap. We attribute this to the extreme density of the NY graph: when constructing the graph using the region-adaptive Delaunay Triangulation (DT) method, the resulting cluster representations tend to become homogenized, reducing discriminative capacity after POI encoding.

However, looking at the $R^2$ coefficients of determination, we observe that for the states of Wyoming and Vermont all competing approaches, including PlaceRep, yield negative $R^2$ values. That means that the model explains less variance (has a higher RMSE) than a naive approach that simply predicts the average population density across all ZIP-code areas in the respective state. This means that the for these states having too few POIs to learn discriminate features from, none of the competitors appear to yield features that can help population density estimate. But this is intuitive, as Wyoming and Vermont are extremely sparsely populated, with Wyoming having an average population density of fewer than 2.5 people per square kilometer and being the second least population U.S. state after Alaska. 
But for the states having larger numbers of POIs and having a larger population density, we observe $R^2$ values close to $1$. For example for Georgia State, with an average population density of 72 people per square kilometer, we observe an $R^2$ of 0.89 indicating that population density variations can be well-explained by PlaceRep features.

\subsubsection{Housing Price Prediction}

Housing prices are a critical indicator of both social well-being and economic vitality, and their prediction has long been a central theme in urban studies, economics, and policy research. Accurate housing price estimation not only informs urban planning and economic forecasting but also guides individual decision-making in residential choices.

In this experiment, we use ZIP-code level embeddings generated by PlaceRep to predict housing prices, ensuring consistency with baseline models. Following prior work, we train a Random Forest regressor with 100 decision trees using an 80/20 train-test split over ZIP-code embeddings. Housing price labels are obtained from the Zillow Research data repository\footnote{\url{https://www.zillow.com/research/data/}}, using the Zillow Home Value Index for August 2024, which provides a standardized measure of typical home values. To evaluate scalability and robustness, we use the same set of states as in the population density task. Each experiment is repeated ten times with randomized splits, and the average performance is reported.

Experimental results, summarized in Table~\ref{tab:main_hp}, demonstrate that PlaceRep consistently outperforms baseline methods across small- and medium-scale datasets. This highlights the effectiveness of our training-free approach in producing expressive region representations that generalize well across prediction tasks. A notable exception is observed in New York, where PlaceRep ranks third with only a marginal gap from the top-performing models. We believe that this gap arises for the same reasons as in the population density prediction task: the POI embeddings become overly homogenized during encoding, which diminishes their discriminative power. 

\begin{figure}[!t]
    \centering
    \includegraphics[width=0.94\columnwidth]{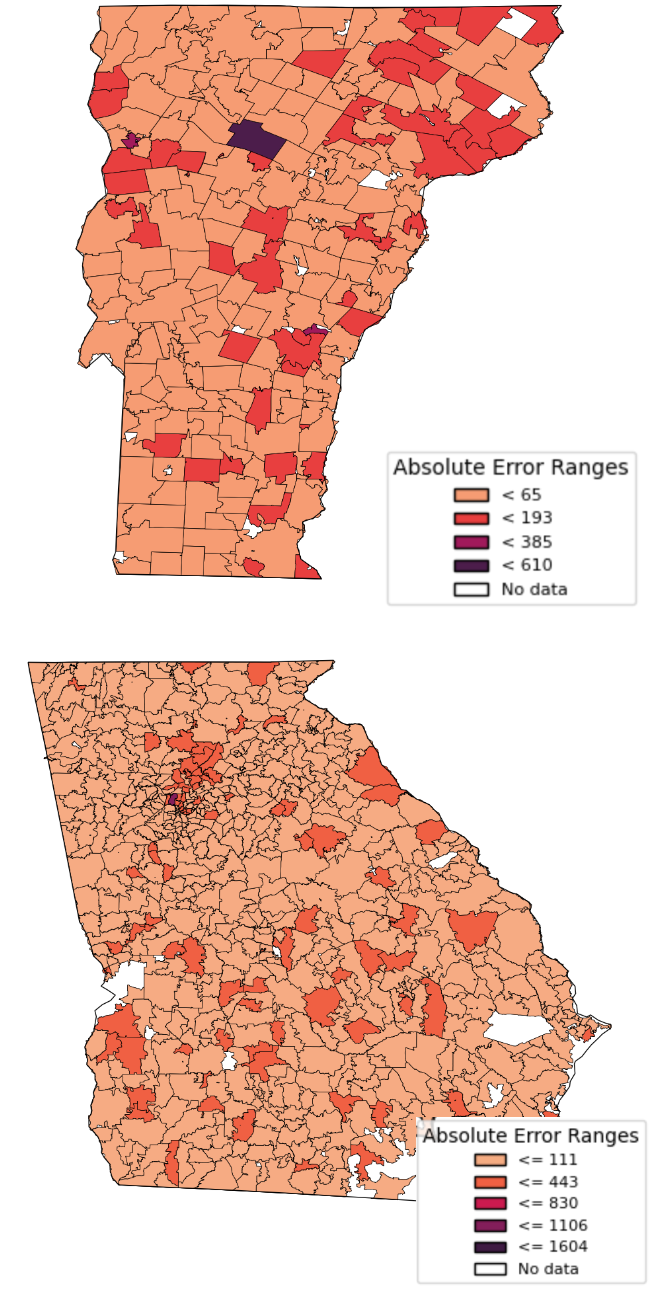}
     \vspace{-0.45cm}
    \caption{Spatial distribution of absolute housing price estimation errors. The top figure shows ZIP-code regions in Vermont (VT), and the bottom figure shows those in Georgia (GA).\vspace{-0.3cm}}\vspace{-0.0cm}
    \label{fig:abs_hp}
\end{figure}

Figure~\ref{fig:abs_hp} illustrates the spatial distribution of absolute errors for Vermont and Georgia. The results show that predictions tend to be more accurate in city centers across both states, while larger discrepancies appear in peripheral and expansive regions. In Georgia, we also observe notable errors in highly dense urban areas, particularly in Atlanta. We attribute these patterns to several factors: \textbf{First,} larger regions tend to be more heterogeneous, making their internal variation harder to capture; \textbf{Second,} highly dense areas may exhibit complex housing market dynamics and sharp local variations that are difficult to model.

While PlaceRep consistently achieves the lowest prediction errors across most states, the negative $R^2$ values in Table~\ref{tab:main_hp} indicate that housing price prediction remains challenging when relying solely on POI-derived representations. Unlike population density, housing prices are influenced by many factors beyond POI information, such as property characteristics and neighborhood conditions. Consequently, POI embeddings alone are insufficient to fully explain housing market variation. Nevertheless, PlaceRep consistently achieves lower RMSE and MAE values than competing approaches, demonstrating that it captures more informative location-based signals than existing place representation methods.

\begin{figure}[!t]
    \centering
    \includegraphics[width=\columnwidth]{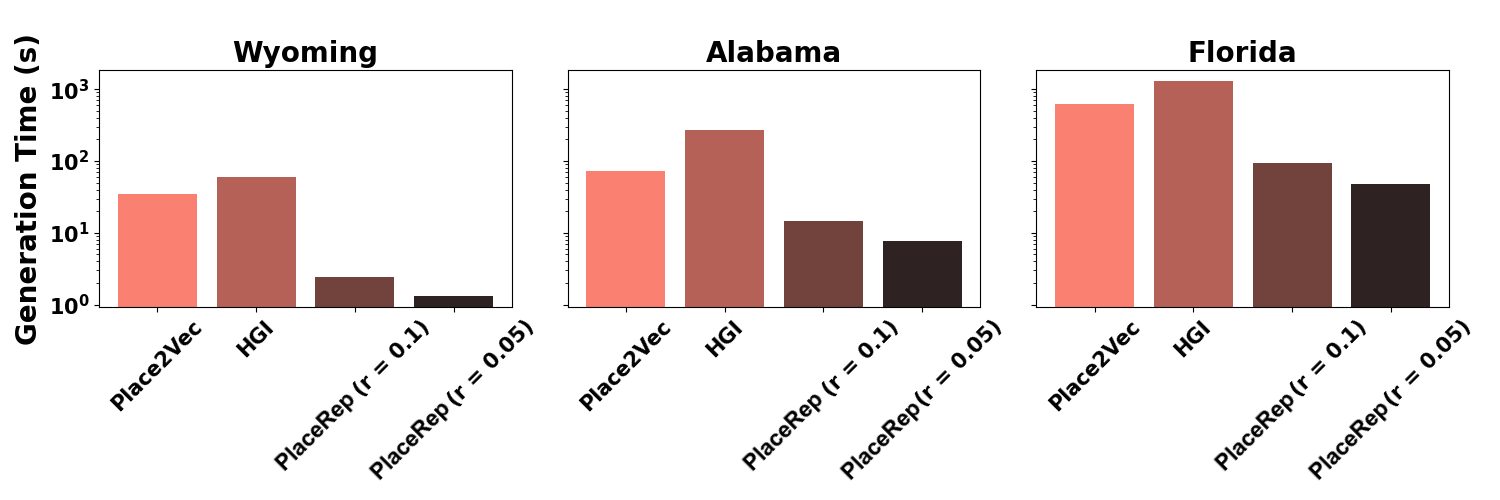}
    \vspace{-0.4cm}  
    \caption{Efficiency comparison of region embedding generation.}
     \vspace{-0.5cm}  
    \label{fig:efficiency}
\end{figure}

\begin{figure*}[!t]
    \centering
    \includegraphics[width=1.0\textwidth]{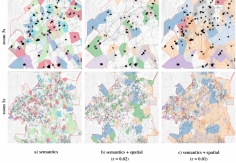}
     \vspace{-0.9cm}
    \caption{Voronoi spatial distribution of identified places in ZIP code 30329, Atlanta, GA. Places with the same color belong to the same cluster. For visualization clarity, only the top 10 identified places with the largest number of POIs are shown in each column. (a) Clustering using only category features. (b) and (c) Clustering using propagated category features.  \vspace{-0.1cm}}

    \label{fig:places}
\end{figure*}

\subsection{Efficiency Comparison}
\label{sec:efficiency}

Figure~\ref{fig:efficiency} illustrates the efficiency and scalability advantages of PlaceRep, which primarily arise from its training-free embedding generation mechanism. For fairness, we exclude Averaging method and PDFM~\citep{agarwal2024general} from this comparison: Averaging incurs negligible computational cost as it simply computes the mean embedding for each region, whereas PDFM relies exclusively on pretrained embeddings from its original implementation. For all other methods, hyperparameters were selected to optimize performance on the downstream population density prediction task described in Section~\ref{sec:downstream_tasks}. For HGI~\citep{huang2023learning}, the number of training epochs was fixed at 100, fewer than those used in the original study.

As summarized in Figure~\ref{fig:efficiency}, PlaceRep demonstrates consistent performance while efficiently managing computational resources even as graph size increases. For instance, in Wyoming, the embedding generation time for PlaceRep is 2.41 seconds for \(r=0.1\) and 1.32 seconds for \(r=0.05\), compared to 34.52 seconds for Place2Vec~\citep{yan2017itdl} and 60.61 seconds for HGI. Similarly, in Alabama, PlaceRep requires 14.61 seconds (\(r=0.1\)) and 7.81 seconds (\(r=0.05\)), while Place2Vec and HGI take 72.12 and 271.24 seconds, respectively. In Florida, the differences are even more pronounced, with PlaceRep completing in 94.65 seconds (\(r=0.1\)) and 48.29 seconds (\(r=0.05\)), versus 621.17 seconds for Place2Vec and 1293.84 seconds for HGI. These results indicate that PlaceRep achieves over an order of magnitude faster embedding generation while its runtime grows much more slowly with graph size. 

Figure~\ref{fig:efficiency} also shows that PlaceRep achieves over an order-of-magnitude speedup in embedding generation compared to baseline methods, with runtime scaling more slowly as graph size increases.

\begin{table}[t]
\vspace{-0.3cm}
\centering
\caption{Optimal hyperparameter settings for each dataset. For each $\alpha_i$, the value shown to the left of the $/$ corresponds to the Population Density prediction task, while the value on the right corresponds to the Housing Price prediction task.\vspace{-0.3cm}}

\label{tab:hyperparam}
\resizebox{\linewidth}{!}{
\begin{tabular}{lccccccc}
\toprule
\textbf{POI Graph} & Reduction Ratio $r$ & $\alpha_0$ & $\alpha_1$ & $\alpha_2$ \\ 
\midrule
\textit{Wyoming (WY)}      & 0.1  &  $0.5$/0.5 & $0.5/0.0$ & $0.0/0,0$ \\
\textit{Vermont (VT)}          & 0.1  &  $0.0/0.0$ & $0.5/0.25$ & $0.0/0.0$ \\
\midrule
\textit{Alabama (AL)}        & 0.1  &  $0.0/0.0$ & $0.0/0.25$ & $0.5/0.0$ \\
\textit{Georgia (GA)}        & 0.05 &  $0.5/0.5$ & $1.0/1.0$ & $0.0/0.0$ \\ 
\midrule
\textit{New York (NY)}    & 0.02 &  $0.25/0.75$ & $0.0/0.0$ & $0.0/0.0$ \\
\textit{Florida (FL)} & 0.05  &  $0.0/0.0$ & $0.5/0.5$ & $1.0/0.0$ \\
\textit{California (CA)}        & 0.05  &  $0.0/1.0$ & $0.5/0.5$ & $1.0/1.0$ \\
\bottomrule
\end{tabular}}
\vspace{-0.3cm}
\end{table}


\subsection{Multi-granular \emph{Place} Identification}

PlaceRep aims to automatically identify \emph{places} within each urban region, defined as subgraphs of POIs or higher-level POI clusters that are both spatially and semantically similar. This enables the discovery of functionally coherent areas within regions such as ZIP codes or cities, supporting applications like place recommendation and urban analytics. For example, clusters of restaurants, cafés, and entertainment venues can reveal vibrant social districts, while clusters dominated by schools, libraries, and parks indicate family-friendly neighborhoods, providing valuable insights for users and urban planners~\citep{islam2022survey,zhao2016survey}.

\begin{table*}[t]\label{tab:transfer}
\centering
\caption{Transferability: ZIP-code embeddings from PlaceRep with different regressor architectures. Units are RMSE, indicating the error of population density estimations (people/km$^2$); lower is better. Best results are in bold.}

\label{tabs:transfer}
\resizebox{0.88\linewidth}{!}{
\begin{tabular}{lccccccccc}
\toprule
\multirow{2}{*}{\bfseries Model} & \multicolumn{3}{c}{\textbf{Vermont (VT)}} & \multicolumn{3}{c}{\textbf{Georgia (GA)}} & \multicolumn{3}{c}{\textbf{Florida (FL)}} \\
\cmidrule(lr){2-4} \cmidrule(lr){5-7} \cmidrule(lr){8-10}
                                        & RF         & MLP       & XGB       & RF          & MLP    & XGB    & RF           & MLP    & XGB \\
\midrule
Averaging                               & 265.38   & 321.45  & 315.64  & 532.00    & 612.11  & 576.42  & 1035.25    & 991.21  & 1141.67  \\
Place2Vec~\citep{yan2017itdl}           & 240.76    & 246.20  & 242.12  & 511.21    & 498.23  & 488.25  & 1115.62    & 892.45  & 976.27  \\
HGI~\citep{huang2023learning}           & 170.94    & 182.55  & 175.98  & 458.21    & 423.05  & \textbf{413.01}  & 772.30    & 762.43  & 730.67  \\
PDFM~\citep{agarwal2024general}         & 308.18    & 292.11  & 254.21  & 536.34    & 500.07  & 478.68  & 806.30    & 714.09  & 726.89  \\
PlaceRep (ours)                          & \textbf{165.23}    & \textbf{171.44}  & \textbf{169.12}  & \textbf{410.16}    & \textbf{401.71}  & 415.12  & \textbf{702.56}    & \textbf{710.50}  & \textbf{724.00}  \\
\bottomrule
\end{tabular}}
\end{table*}

Our clustering-based approach reduces the large-scale POI graph of each region into a predefined number of places, given by $k_r = \left\lfloor n_r \times r \right\rfloor$, where $n_r$ is the number of POIs in region $r$ and $r$ is the reduction ratio. The parameter $r$ determines the granularity of the identified places. Larger values of $r$ yield more fine-grained clusters, capturing smaller and more specialized neighborhoods, whereas smaller values of $r$ lead to more general and aggregated places that encompass a wider variety of POI types. In this way, $r$ serves as a natural control parameter for the level of granularity at which urban functionality is represented.

Figure~\ref{fig:places} illustrates this process using Voronoi diagrams~\citep{aurenhammer1996voronoi}, which depict the nearest-area partitioning around each POI in ZIP code 30329, Atlanta, GA, after the place generation process. POIs sharing the same Voronoi color are clustered into the same place. The upper row provides a $3\times$ zoomed view of the lower row to highlight POIs in greater detail. For visualization purposes, in all three cases we display only the top 10 identified places with the largest number of POIs after clustering:

\textbf{Semantic-only-based Places.} In column (a), representations of places are obtained by clustering solely on semantic information, using only category-level POI features. This results in discontinuous and fragmented clusters, where each color corresponds to a semantic category of POIs. This approach overlooks the influence of neighboring POIs in the vicinity of each POI, thereby ignoring valuable spatial context when constructing a representation for a place.

\textbf{Semantic- and Spatial-Based Places.}
Unlike column (a), columns (b) and (c) leverage both semantic and spatial feature aggregation, producing spatially contiguous and semantically coherent clusters that better reflect meaningful urban places. This property is critical for traceability: \textbf{First,} each place embedding can be directly linked to its underlying POIs, including their spatial configuration and semantic similarity, enabling interpretable representations and explaining functional similarities or differences between regions. \textbf{Second,} aggregating POIs into coherent clusters provides a structured view of the urban landscape, making the relationship between individual places and the overall region explicit and supporting a richer understanding of regional functional organization.
 
In column (c), places highlighted in red contain mostly POIs from \emph{Dining and Drinking}, with some from \emph{Business and Professional Services}, indicating mixed-use clusters where restaurants, cafés, and bars co-locate with offices and service-oriented businesses. Such patterns commonly occur in downtown districts or commercial corridors, where dining venues are interwoven with professional offices, coworking spaces, and service providers. 

\textbf{Effect of reduction ratio $r$.} As the reduction ratio $r$ increases, more clusters emerge, capturing finer-grained structures, while smaller values of $r$ highlight broader, more general areas. In column (b), compared to column (c), identified places appear more discontinuous and fragmented; however, each place highlights semantically and spatially coherent POIs in greater detail. For instance, in column (b), the green Voronoi areas contain POIs predominantly from category level~1 of \emph{Community and Government} and \emph{Business and Professional Services}, which may represent local government offices, community centers, libraries, as well as law firms or consulting offices. Such detailed, specialized clusters are less apparent in the more general, aggregated places shown in column (c), demonstrating how varying $r$ allows control over the granularity of identified urban places and their functional specificity.

\subsection{Transferability Comparison}

An essential criterion for assessing urban region representations is their ability to support diverse downstream architectures from a data-centric perspective. Unlike PlaceRep, which produces embeddings through a training-free mechanism, most existing approaches rely on backbone GNNs or other types of neural network architectures for embedding generation. This dependency can introduce inductive biases that limit their flexibility when applied to different downstream models.

Table~\ref{tabs:transfer} presents results demonstrating that embeddings derived from PlaceRep generalize robustly across three regression architectures: Random Forest (RF), Multi-Layer Perceptron (MLP, with two hidden layers of 32 and 16 neurons), and Extreme Gradient Boosting (XGB). With the exception of a single case in Georgia, where XGB trained on HGI embeddings achieves the best performance, PlaceRep consistently outperforms competing baselines. Reported values reflect the mean performance over ten runs of downstream inference, following the same setup as Section~\ref{sec:downstream_tasks}, with optimal hyperparameters selected for each model.

\section{Conclusions}
\label{sec:conclusion}
We introduced PlaceRep, a training-free model that generates general-purpose place embeddings by capturing spatial context and neighborhood structure. Our comprehensive experiments demonstrate its effectiveness in producing region-level embeddings at multiple geographic scales, which can be readily applied to a variety of urban downstream tasks. In this study, the only data modality used for POI features was category information. As future work, PlaceRep could be extended to leverage multi-modal data, including mobility information, to capture human movement patterns, enabling the learning of more informative and meaningful region-level and place-level representations.
\clearpage

\bibliographystyle{ACM-Reference-Format}
\bibliography{main}

\end{document}